\documentclass[manuscript,screen]{acmart}

\AtBeginDocument{%
  \providecommand\BibTeX{{%
    \normalfont B\kern-0.5em{\scshape i\kern-0.25em b}\kern-0.8em\TeX}}}

\setcopyright{acmcopyright}
\copyrightyear{2021}
\acmYear{2021}
\acmDOI{10.1145/1122445.1122456}

\acmConference[Smell]{Smell, Taste, and Temperature Interfaces}{May 7--9, 2021}{Online}
\acmBooktitle{Smell, Taste, and Temperature Interfaces}
\acmPrice{15.00}
\acmISBN{978-1-4503-XXXX-X/21/06}

\usepackage[utf8]{inputenc}

\begin{document}

\title{Fragmented and Valuable: Following Sentiment Changes in Food Tweets}


\author{Maija Kāle}
\authornote{Both authors contributed equally to this research.}
\authornotemark[1]
\email{maijakale@gmail.com}
\affiliation{%
  \institution{Faculty of Computing, University of Latvia}
  \city{Riga}
  \country{Latvia}
}

\author{Matīss Rikters}
\email{matiss@logos.t.u-tokyo.ac.jp}
\orcid{0000-0002-3530-6873}
\affiliation{%
  \institution{The University of Tokyo}
  \city{Tokyo}
  \country{Japan}}
\email{xxx}

\renewcommand{\shortauthors}{Kāle and Rikters}

\begin{abstract}
  We analysed sentiment and frequencies related to smell, taste and temperature expressed by food tweets in the Latvian language. To get a better understanding of the role of smell, taste and temperature in the mental map of food associations, we looked at such categories as `tasty' and `healthy', which turned out to be mutually exclusive. By analysing the occurrence frequency of words associated with these categories, we discovered that food discourse overall was permeated by `tasty' while the category of `healthy' was relatively small. Finally, we used the analysis of temporal dynamics to see if we can trace seasonality or other temporal aspects in smell, taste and temperature as reflected in food tweets. Understanding the composition of social media content with relation to smell, taste and temperature in food tweets allows us to develop our work further - on food culture/seasonality and its relation to temperature, on our limited capacity to express smell-related sentiments, and the lack of the paradigm of taste in discussing food healthiness.
\end{abstract}



\maketitle

\section{Introduction}

The human-food relationship can be explored via public health data - such as the increase of obesity, Type 2 diabetes and cardiovascular diseases globally -, as well as by looking at the digital behaviour related to food: food photos, menu descriptions, food blog entries, tweets and posts on other social networks that are occupying an increasingly large share of the digital content and social media. Food is richly documented and discussed in multiple formats in the contemporary media, serving as a valid manifestation of the ‘Zeitgeist’. “Unlike many consumer products (such as electronics), food is fragile, emotional, cultural” \cite{metcalfe2019food}, thus playing a constant part throughout the entire human evolution from hunter-gatherers to super-consumers \cite{article} of today. 

Language and the way we talk about food constitutes a whole research area that contains an abundance of data, albeit not fully comprehended and analysed yet. For example, research by Dan Jurafsky shows the extent to which food descriptions are interlocked within the human language; it also points out certain peculiarities in our ability to describe food experiences - turns out that our expression of negative feelings is much more diverse \cite{jurafsky_language_2014}. Meanwhile, social media networks are generally believed to showcase an overwhelmingly positive experience. This is also supported by research on wine reviews, where negative reviews were found to be substantially less frequent than the positive ones \cite{WineExperts}. We have to admit that our knowledge about the language of food is fragmented and that the utility of big data available from social media platforms is still under-researched when it comes to the subtleties of human and food relationship. An increased grasp of the textual expressions of the human-food relationship can lead to a better understanding of the factors influencing our food choice, which could potentially contribute to a healthier lifestyle.

Categorising various food types into ‘healthy’ and ‘unhealthy’ proved to be challenging. Healthiness as a concept depends on many aspects, far broader than food-related factors alone. Although the perception of healthiness is largely subjective, a general observation is that "the objectives of health and taste often conflict - and taste usually prevails in food decision making” \cite{Maiarticle}.
Words describing the food play a crucial role in positioning that food on the mental map. For instance, characterising the food as healthy has been proven to decrease the perception of its tastiness (unhealthy = tasty intuition \cite{doi:10.1509/jppm.14.006}). 

In this work we analysed the sentiments of food tweets in terms of their positive, neutral and negative valence. We observed the depictions of smell, taste and temperature in relation to `healthy' vs `tasty' foods and performed a temporal dynamics analysis in order to better grasp if and how food-related sentiment has varied over time.

We inspected the Latvian Twitter Eater Corpus (LTEC \cite{SprogisRikters2020BalticHLT}), which had been collected over nine years following 363 eating-related keywords in the Latvian language. Our work has allowed us to capture fragmented yet valuable notions of human and food relationship depicted on social media (in this case Twitter) with regard to smell, taste and temperature. It can serve as the basis for future research, deepening our understanding of the temporal dynamics, the frequency of occurrence as well as interconnections among the various concepts appearing in tweets.

\section{Methodology}

The LTEC has a total size of $\sim$2.4M tweets generated by $\sim$169k users. For the experiments with smell, taste and temperature, we selected only unique tweets containing at least one of the predefined adjectives or verbs (Table \ref{tab:word-table}) in any valid inflection (automatically inflected using a Finite State Morphology Tool for Latvian \cite{deksne-2013-finite}).

We adapted the sentiment analysis tool \footnote{Sentiment Analysis Toolkit - \url{https://github.com/M4t1ss/sentiment-analysis-toolkit}} used for general Latvian tweets \cite{Thakkar2020pretraining} and used it along with a Latvian-specific language model \cite{viksna2020large}, which we fine-tuned onto the entire LTEC. This allowed us to train a sentiment classifier to distinguish negative, positive and neutral tweets in the sentiment training data subset of LTEC other Latvian tweet data sets\footnote{LV Twitter sentiment corpus - \url{https://github.com/nicemanis/LV-twitter-sentiment-corpus}} \cite{pinnis2018latvian,viksna2018sentiment,peisenieksuses}. In total we had 17,255 tweets for training, 1,000 \cite{pinnis2018latvian} for development and 744 from the LTEC for evaluation. The precision of our best classifier on the evaluation set reached 74.06\%, which is close to human performance.

\begin{table}[b]
    \begin{tabular}{|l|lllllll|}
    \hline
    \textbf{Smell}       & Smell     & Smelly & Stink    & Stinky &         &       &      \\ \hline
    \textbf{Taste}       & Delicious & Tasty  & Not tasty  & Bitter & Sweet   & Salty & Sour \\ \hline
    \textbf{Temperature} & Cold      & Hot    & Lukewarm & Warm   & Boiling & Cool  &      \\ \hline
    \end{tabular}
    \caption{Verbs and adjectives describing smell, taste and temperature.}
    \label{tab:word-table}
\end{table}

\section{Results}

\begin{figure}[b]
  \includegraphics[width=\linewidth]{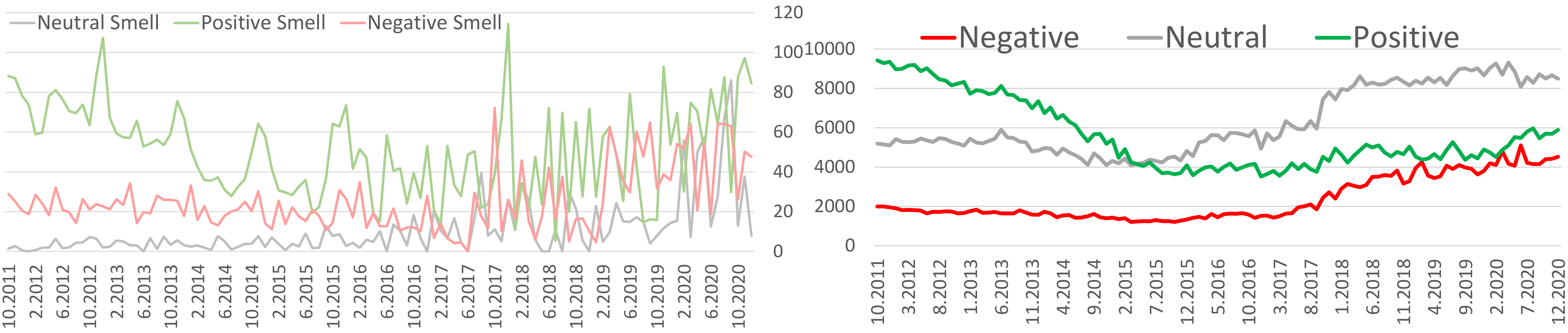}
  \caption{Distribution of tweet sentiment in total on the right and sentiment related to smell on the left over time from 2011 to 2020.}
  \label{fig:sentiment-time}
\end{figure}

Figure \ref{fig:sentiment-time} (right) shows how sentiment in food/eating-related tweets has changed from 2011 to 2020. We can see a gradual decrease in positive sentiment and a steady rise of negative and neutral sentiment. This might have been caused by the rapid expansion of Instagram and other social networks containing an abundance of positive food posts, which has subsequently relegated Twitter's status in this matter to a `book of complaints'. With these data we can challenge the assumption that social media (more specifically, Twitter) are a platform for sharing predominantly positive experiences, as was previously indicated by the research on wine reviews \cite{WineExperts}. The Twitter data contain a large volume of neutral tweets (potentially created by food bots, e.g. lunch offers, general information on food) and exhibit a growing tendency towards an equal share of positive and negative tweets.

\begin{figure}[b]
  \includegraphics[width=\linewidth]{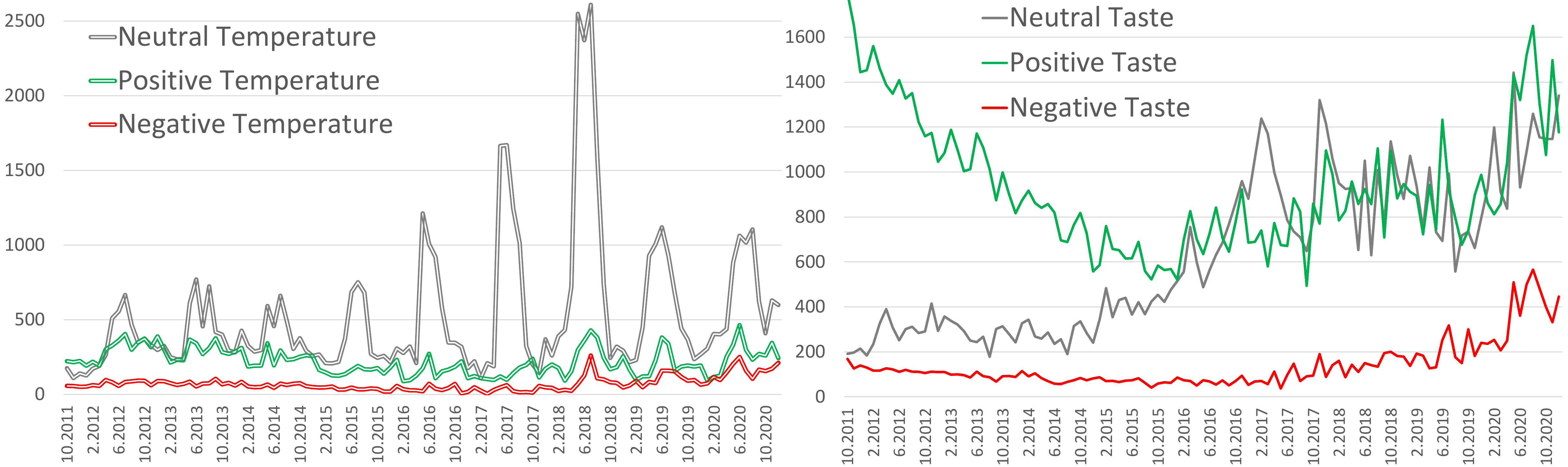}
  \caption{Temporal dynamics of taste and temperature sentiments in tweets, 2011-2020.}
  \label{fig:sentiment-taste-temperature}
\end{figure}

Figure \ref{fig:sentiment-taste-temperature} shows the sentiment changes in relation to taste and temperature; Figure \ref{fig:sentiment-time} (left) - in relation to smell, which had by far the lowest number of tweet entries in the data set. Concerning temperature-related sentiment patterns, we can observe periodically repetitive entries containing positive, negative, and, in particular, neutral sentiment occurring during the summer months. This could be attributed to one particular food - cold beet soup, which is widely consumed in Latvia and the Baltic States in summer. Overall, temperature does not appear to cause much polarity in sentiment - it is mostly perceived as neutral. The category of taste, however, is different. It yields an overwhelmingly positive sentiment, although with a sharp increase of neutral and a gradual rise of negative sentiment. Smell, on the other hand, seems to be the opposite of the category of temperature. Although appearing relatively seldom, it is associated with either positive or negative sentiments, but rarely so with neutrality.

Table \ref{tab:tasty-healthy} shows how often smell, taste and temperature words are collocated with the adjectives `tasty' or `healthy' in a tweet. The results allow us to conclude that in the food discourse `tasty' takes precedence over `healthy'. It supports the notion that Twitter is used for the expression of immediate emotions, where hedonistic (as well as negative) representations of food consumption outnumbers considered and healthy connotations. This impulsive and instant nature of tweeting is also reflected in seasonal variations - for instance, in the case of the cold soup during the summer months. It proves that tweets can provide important information on the emotional response to food in real time, which could be useful for e.g. personalised health apps and their data analysis.

\begin{table}[t]
    \begin{tabular}{|l|c|c|c|l|c|c|c|}
    \hline
            & Taste & Temperature & Smell & & Taste & Temperature & Smell\\ \hline
    \textbf{Tasty}   & 3180  & 3619        & 860   &  \textbf{Healthy} & 293   & 305         & 23 
     \\ \hline
    \end{tabular}
    \caption{The number of smell, taste and temperature words collocated with tasty or healthy.}
    \label{tab:tasty-healthy}
\end{table}

\section{Conclusion}

The language concerning food that is used on social media can potentially influence our food preferences. Words describing food can affect the categorisation of food into `healthy' or `tasty', which according to theory are mutually exclusive categories \cite{Maiarticle}. In our work we analysed the language of food on Twitter in Latvian, tracing the general sentiments as well as looking into the particulars related to the categories of smell, taste and temperature. We also examined occurrence frequencies of smell, taste and temperature words in relation to `tasty' or `healthy' foods.

Our results indicate that food tweets in Latvian are not dominated by positive sentiments. Instead we could observe a recent increase in neutral tweets and a similar share of positively versus negatively coloured tweets. This has changed over time, as several years ago positive sentiments were prevalent. 

There are substantial differences in sentiment variations related to smell, taste and temperature. Temperature is the most `neutral' one; taste and smell tend to be mostly positive or negative, of which taste has yielded the highest share of positive tweets. Regarding the `healthy' versus `tasty' dichotomy, the majority of tweets contain associations with `tasty' whereas connotations with `healthy' are scarce.
We can hereby conclude that large-scale social network data can improve our understanding of the human-food relationship and help us develop strategies and tactics for instilling healthier - although not necessarily less tasty - food habits. By researching on food-related behaviour on social media we can move from fragmented and valuable data to a more comprehensive awareness of the food choice mechanisms and their associated sentiments.

\bibliographystyle{ACM-Reference-Format}
\bibliography{acmart}

\end{document}